\documentclass[11pt]{article}
\usepackage[a4paper, margin=1in]{geometry}

\usepackage[T1]{fontenc}
\usepackage{graphicx}
\usepackage{multicol}
\usepackage{keyval}
\usepackage{url}
\usepackage{natbib}
\usepackage{booktabs}
\usepackage{amsmath,amssymb}
\usepackage[svgnames]{xcolor}
\definecolor{linkblue}{rgb}{0, 0.1, 0.7}
\usepackage[colorlinks,linkcolor=linkblue,urlcolor=linkblue,citecolor=linkblue]{hyperref}
\usepackage{cleveref}
\usepackage{enumitem}
\usepackage{colortbl}
\usepackage[bottom]{footmisc} % footnotes at the bottom of the page

\graphicspath{{img/}}

\begin{document}

% \fontfamily{utopia}\selectfont
\fontfamily{stix}\selectfont

\title{
Fusing Sentence Embeddings Into \\ LSTM-based Autoregressive Language Models
}

\author{
    \textbf{Vilém Zouhar} \qquad \textbf{Marius Mosbach} \qquad \textbf{Dietrich Klakow} \\
    % \hspace{6mm}\begin{minipage}[c]{10cm}
    %     \centering
        Universität des Saarlandes, Saarland Informatics Campus,\\
        Sprach- und Signalverarbeitung (LSV) \\
    % \end{minipage}
    % \hspace{0.1mm}
    % \begin{minipage}[c]{6mm}
    %     \includegraphics[width=\linewidth]{img/lsv.png}
    % \end{minipage} \\
    \texttt{\{vzouhar,mmosbach,dietrich.klakow\}@lsv.uni-saarland.de}
}

\date{}

\maketitle

\begin{abstract}
Although masked language models are highly performant and widely adopted by NLP practitioners, they can not be easily used for autoregressive language modelling (next word prediction and sequence probability estimation). 
We present an LSTM-based autoregressive language model which uses prefix embeddings (from a pretrained masked language model) via fusion (e.g. concatenation) to obtain a richer context representation for language modelling.
We find that fusion helps reliably in lowering the perplexity ($16.74 \rightarrow 15.80$), which is even preserved after a transfer to a dataset from a different domain than the training data.
We also evaluate the best-performing fusion model by correlating its next word surprisal estimates with human reading times. 
Contradicting our expectation, and despite the improvement in perplexity overall, the correlation remains the same as for the baseline model.
Lastly, while we focus on language models pre-trained on text as the sources for the fusion, our approach can be possibly extended to fuse any information represented as a fixed-size vector into an auto-regressive language model.
These include e.g. sentence external information retrieved for a knowledge base or representations of multi-modal encoders.

\vspace{4mm}
\hfill
\begin{minipage}[c]{6mm}
\includegraphics[width=\linewidth]{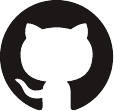}
\end{minipage}
\hspace{1mm}
\begin{minipage}[c]{7.4cm}
\href{https://github.com/zouharvi/sentence-embd-fusion}{\texttt{github.com/zouharvi/sentence-embd-fusion}}
\end{minipage}
\end{abstract}

% set previous font
\fontfamily{lmodern}\selectfont

% Storyline inspiration
% https://arxiv.org/abs/2005.09471#:~:text=Our%20analysis%20shows%20Transformers%20to,evidence%20for%20cue%2Dbased%20retrieval.
% https://repository.ubn.ru.nl/bitstream/handle/2066/213724/213724.pdf?sequence=1
% https://arxiv.org/abs/2006.01912

\section{Introduction}

The task of masked language modelling is to predict a token (or multiple) under a mask in a sequence:
$p(w_i| w_1 \ldots, w_{i-1}, w_{i+1}, \ldots w_{|s|})$.
Masked language modelling as a training objective is inspired by the cloze task \citep{taylor1953cloze, devlin2018bert} and has recently led to impressive improvements in various natural language tasks \citep{zaib2020short}.
In contrast, autoregressive LMs model the ``sequence probability'' $p(w_1, \ldots, w_{|s|})$ as $\prod_{i=1}^{|s|} p(w_i|w_{<i})$.
In practice, this allows the model to be used for decoding and generating sequences.
The equivalent in human experiments is the word-level Shannon game \citep{shannon1951prediction} (predicting the next word based on the left context).

Autoregressive LMs are used in a plethora of places.
The most intuitive and straightforward application is text auto-completion \citep{jaech2018personalized}.
Moreover, the ability of the model to compute sequence probabilities and compare them is highly useful in many NLP applications which make use of sequences of text either as an intermediate or final output.
These include machine translation \citep{brants2007large,gulcehre2017integrating}, speech recognition \citep{toshniwal2018comparison} or free-form question answering \citep{dong2019unified}.
A commonly used part of these pipelines is the decoding for which an autoregressive LM is essential \citep{chorowski2016towards}.

On the other hand, masked LM has shown to be a very efficient training task that yields encoder models which became widely used in the NLP community \citep{xia2020bert,koroteev2021bert}.
One commonality of both autoregressive and masked LMs is that the context representations they are forming for a given input, either a representation of the previous words in the sequence (autoregressive) or representations of the whole input sequence except the mask part (masked LM), depend only on the training data of the model as well as its weights.
It might however be desirable to ``enrich'' this representation with information external to the model. 

\begin{figure}[htbp]
\centering
\includegraphics[width=0.75\linewidth]{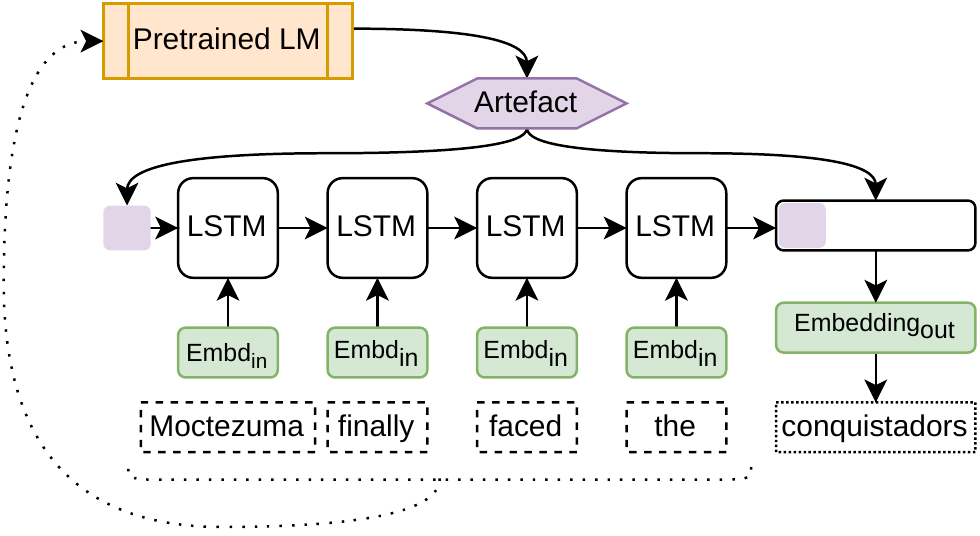}
\caption{Example architecture of an LSTM language model enhanced with fusion (early and late) from a pretrained LM illustrated on a sample sentence. Knowledge of \underline{Moctezuma} helps in the next word prediction.}
\label{fig:model_moctezuma}
\end{figure}

\citet{zouhar2022artefact} present a conceptual framework in which items retrieved from a persistent storage, dubbed artefacts, are fused into a knowledge-intensive model to improve its predictions.
This can be applied to a range of tasks: prototypically question answering and fact verification but to others as well, such as knowledgable dialogue or LM.
We take this line of thought further and use fusion to leverage both the strength of a pre-trained masked LM: rich context representations for producing the artefacts, with the properties of an autoregressive LM: easy decoding.
Below we briefly summarize our main contributions:

\begin{itemize}
\item Design of a simple autoregressive LM with sentence embedding fusion.
\item Comparison of various fusion modes.
\item Exploration of the dependency of the model on the artefact quality.
\item Evaluation of the models via correlation with human surprisal.
\item Experiments with fusion to help with cross-domain performance.
\end{itemize}

% The code to replicate the experiments in this paper is available open-source.\footnote{\href{https://github.com/zouharvi/sentence-embd-fusion}{github.com/zouharvi/sentence-embd-fusion}}

\section{Related Work}

The Transformer architecture \citep{vaswani2017attention}, originally encoder-decoder and proposed for machine translation, is ubiquitous in modern NLP \citep{wolf-etal-2020-transformers,kalyan2021ammus,lin2021survey}.
Follow-up work produced encoder-only models, such as BERT \citep{devlin2018bert}, RoBERTa \citep{liu2019roberta} or SentenceBert \citep{reimers2019sentence}, which were successful in providing dense representations of the input text \citep{liu2020representation} and tackling a variety of challenging natural language understanding tasks \citep{wang2018glue,wang2019superglue}.
Opposed to that are decoder-only models, such as GPT \citep{radfordimproving} and its variations \citep{radfordlanguage,brownlanguage,gpt-j}.
They are capable of autoregressive LM (i.e. prediction of the next word's distribution).

\paragraph{MLM.}

% TODO: fix non-autoregressive != not autoregressive/MLM

There have been multiple approaches to enhancing masked LMs to gain some properties of autoregressive LMs, namely word/sentence probability scoring and iterative text generation.
\citet{liao-etal-2020-probabilistically} propose a probabilistically masked LM capable of text generation.
Text generation can also be done with masked-language models, e.g. BERT \citep{su2021non,jiang2021improving}.
Apart from left to right text generation, estimating sentence/word probabilities from masked language models is also useful for various NLP tasks.
\citet{salazar-etal-2020-masked} mask individual words in the sentence independently to compute pseudo-log-likelihood scores.
\citet{wang2019bert} incorporate BERT into Markov Random Field structure which also makes it possible to evaluate sequence probability.

\paragraph{Fusion.}
\citet{khandelwal2019generalization} proposed an LM which precomputed embeddings of LM contexts + correct following word from the training data.
During inference, the model would compute the context which would be used for retrieval and the next word would be determined by both the model's actual prediction and the retrieved next word (convex combination of the two distributions).
This would be considered a super-late fusion because the artefact is not used in any computation and the combination weight is a hyperparameter.
This was improved by \citet{yogatama2021adaptive} by computing the weights dynamically based on the current context.
This allows the model to use its own predictions for easy and predictable words, such as articles and to fall back to retrieval for words which require memorization, such as named entities.
Named entities are the focus of \citet{logan2019barack}.
They build a local knowledge graph which is used by the model for the prediction of named entities.

\paragraph{Multimodality.}
The artefacts used need not be limited to vectors or text.
Multimodal settings, such as images or video, have also been used successfully to leverage the extra information \citep{anand2021multimodal,mogadala2021trends}.

\paragraph{Memory augmentation.}
Fusion is a specific mechanism used in memory augmented models.
\citet{zhong2022training} distinguish between local memory, long-term memory and external memory and uses them in prediction via super-late fusion.
The pointer-sentinel mixture model by \citet{merity2016pointer} computes a distribution based on all the input words so far and combines it using a gate.
A similar approach to using recent history is also used by \citet{grave2016improving} in the continuous cache.

\section{Methods}

% \TODO{we call the output artefact as well}

\subsection{Data}

We use the WikiText-103 \citep{merity2016pointer} corpus for training and evaluation.
Additionally, we make use of BookCorpus \citep{Zhu_2015_ICCV} and CC-News \citep{cc_news} for evaluation of cross-domain performance.
Natural Stories \citep{futrell2021natural} are used solely for evaluation of reading time correlation.
Although modern large LMs are trained on corpora with 1-100M sentences \citep{kaplan2020scaling,martin2020camembert}, this work is limited by available computational resources.
It is partly justified by using an order of less parameters (13M for our LSTM-based LM, 110M for BERT-base) and \Cref{subsec:limited_data} which explores the effect of limited training data on fusion behaviour.

\begin{table}[htbp]
\center
\begin{tabular}{lcc}
\toprule
& \textbf{Train} & \textbf{Dev/Eval} \\
\midrule
WikiText-103 & 100k & 10k \\
BookCorpus & $-$ & 10k \\
CC-News & $-$ & 10k \\
Natural Stories & $-$ & 430 \\
\bottomrule
\end{tabular}
\caption{Number of sentences in splits in used datasets.}
\label{tab:data_size}
\end{table}

\subsection{Model}
\label{subsec:model}

Our autoregressive language model is intentionally small and composed of: (1) input embedding layer which maps from the vocabulary to 512 dimensions, (2) LSTM with hidden state dimensionality of 256, (3) linear projection layer which maps back to 512 dimensions, (4) output embedding layer to 8192 dimensions (the size of our vocabulary).
The artefact vector is provided by a pretrained MLM which is not fine-tuned and described in greater detail in \Cref{subsec:artefact_source}.
Batches are constructed from individual sentences, i.e. a sentence of 10 words creates a batch of 11 examples for next word prediction (the first input it \texttt{<BOS>} and the last one the whole sentence based on which \texttt{<EOS>} should be predicted).

The fusion happens either by the initial LSTM hidden state being the artefact (in 2) or by either concatenation or an arithmetic operation with the last hidden state of the LSTM (in 3).
Unless specified otherwise, we concatenate the representation of the BERT$_\textsc{Base-Uncased}$ CLS-token (the artefact) with the last hidden state.
The artefact for the word $w_t$ is computed as $\xi(w_{<t})$, therefore our autoregressive LM is a function $p(w_t|w_{<t}, \xi(w_{<t}))$.
Both early and late fusion use the same artefact.

For experiments in \Cref{subsec:artefact_source} we use three dense layers of 768 dimensionalities separated with ReLU.
All models use the same hyperparameters, and are optimized with the Adam optimizer \citep{kingma2014adam} using a learning rate of $10^{-6}$, a batch size of $|S|$ and a BPE vocabulary of size 8196.
For all experiments, we use Byte-Pair-Encoding \citep{shibata1999byte} with a vocabulary size of 8192.
The perplexity is evaluated on the subword level.

% according to the following set of equations:
% \begin{align*}
% & v_{:i} = \text{One-hot}(s_{:i}) \\
% & x_{:i} = \text{Linear}^{8192\rightarrow 512}(v_{:i}) & \textit{Input embedding layer} \\
% & x'_i = \text{LSTM}^{256}(x)_{[-1]} & \textit{Last hidden state} \\\
% & 
% \end{align*}

% The second reviewer asked for this figure even though it's a bit redundant.
\begin{figure}[htbp]
\centering
\includegraphics[width=0.6\linewidth]{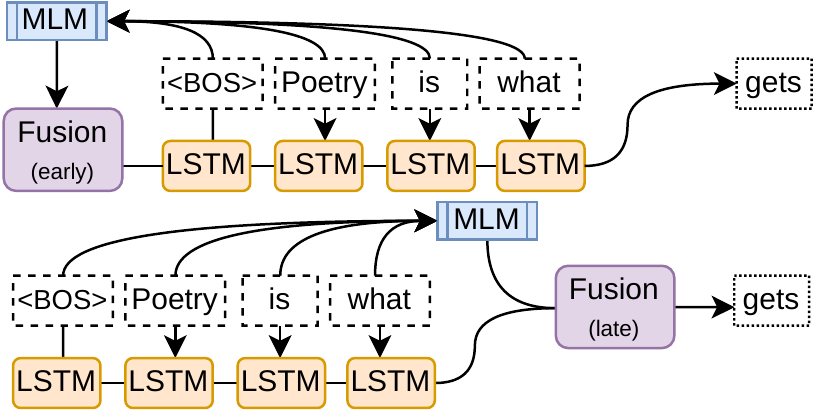}
\caption{In-detail scheme of using MLM to provide artefacts for both early (top) and late (bottom) fusion.\protect\footnotemark}
\label{fig:model_moctezuma_detailed}
\end{figure}

\footnotetext{Original quote by Robert Frost: \emph{Poetry is what gets lost in translation.}}

\section{Experiments}
\label{sec:experiments}

\subsection{Fusion Modes}

We make use of two groups of fusion, based on at which point in the computation the artefact is presented to the model:

\begin{itemize}
\item \textbf{Late fusion}: the sentence embedding vector is fused with the final hidden state vector used for prediction via either (1) concatenation (changes the dimensionality to $768+768$), (2) addition, or (3) multiplication.
\item \textbf{Early fusion}: the sentence embedding vector becomes either (1) the initial hidden state $h_0$ of the LSTM, (2) the initial cell state $c_0$ of the LSTM or (3) both.
\end{itemize}

\begin{figure}[htbp]
\centering
\includegraphics[width=0.9\linewidth]{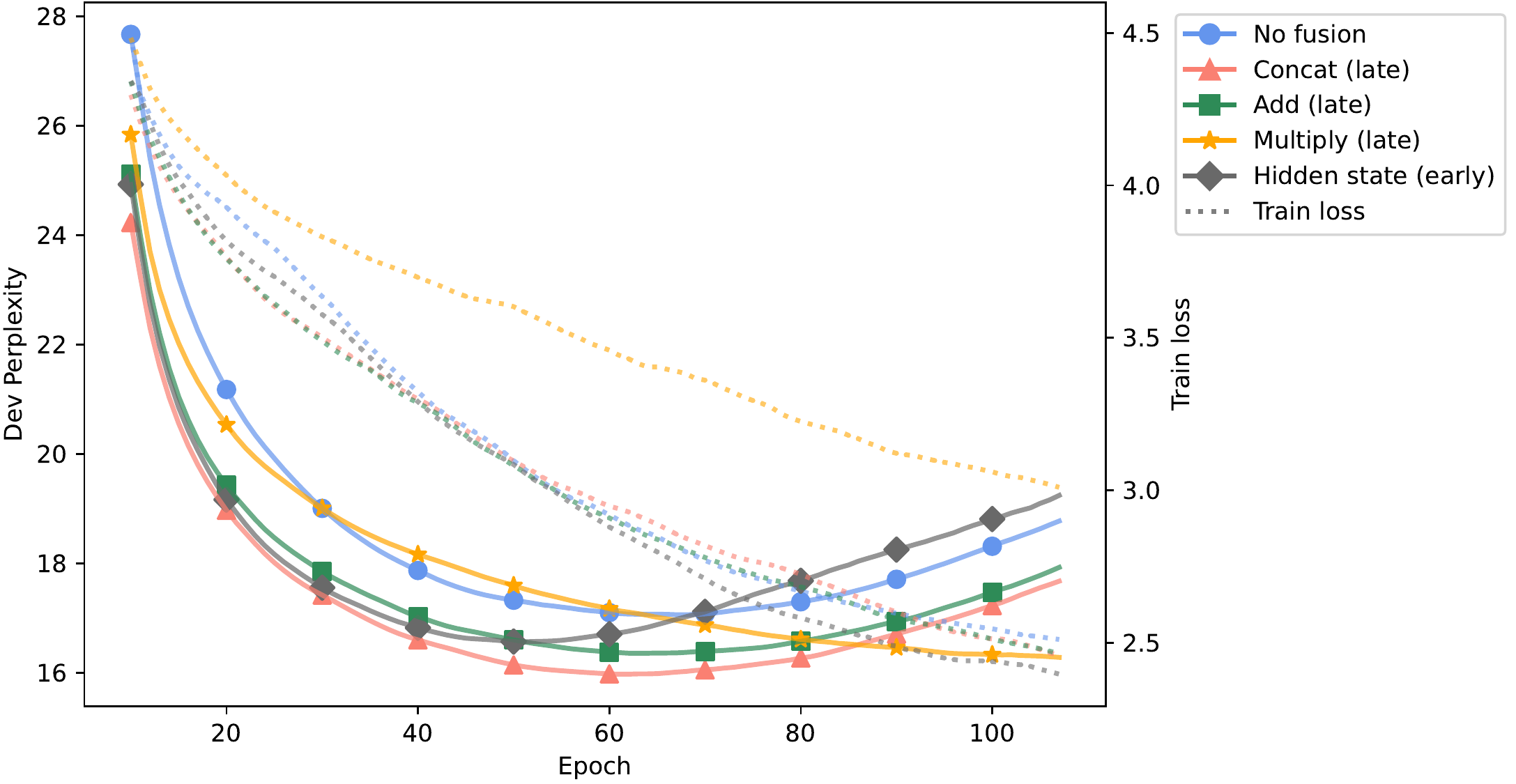}
\caption{Model performance during training of models with and without fusion of sentence embeddings. First epoch cropped for increased resolution.}
\label{fig:fusions}
\end{figure}

The comparison of fusion modes is shown in \Cref{fig:fusions}.
An immediately noticeable difference is that most fusions outperform the model without fusion (blue).
This means that the sentence embedding information is being utilized and helpful for predicting the next word.
Next, the early fusion reaches its minimum earlier (i.e. fewer training steps) than the other fusion methods and the multiplication fusion takes much longer to converge.
The graph is cropped at a point of convergence.
In theory, the sooner the model has access to the artefact, the larger should be its impact on the computation performed by the model.
In practice, sometimes this is not necessary, such as in the nearest-neighbour language models \citep{khandelwal2019generalization} and it may hurt the optimization or the information may get lost.

% TODO: for smaller data early fusion is better - why?

\subsection{Artefact Context Size}

\begin{figure}[htbp]
\centering
\includegraphics[width=0.65\linewidth]{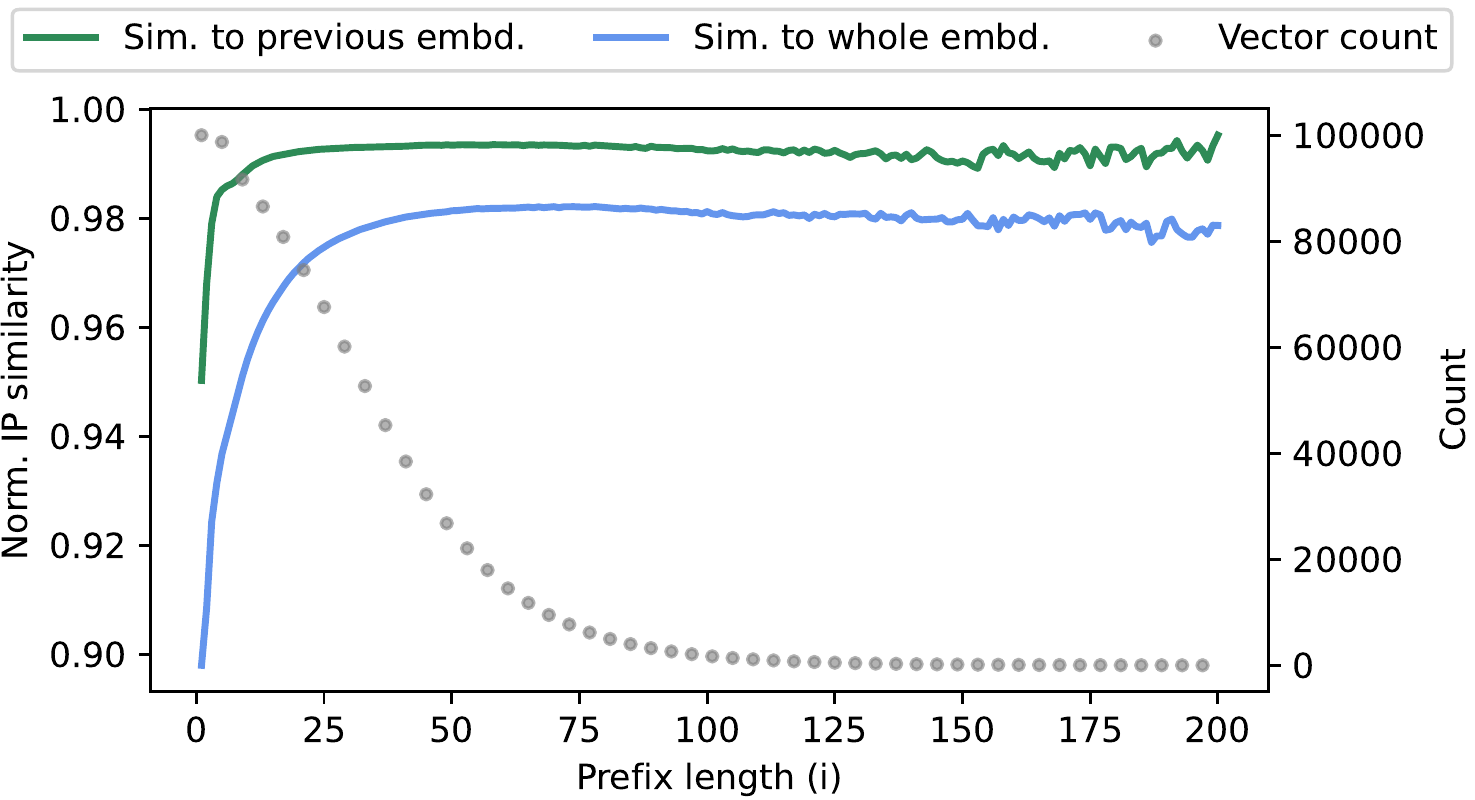}
\caption{Normalized inner product similarity between (1) consecutive vectors and (2) current and last embedding. The values are averaged and the x-axis is cropped for the counts to be at least 10.}
\label{fig:embd_sim}
\end{figure}

Recomputing the sentence prefix embeddings is expensive and needs to be done after every newly generated token.
Because the vectors become more and more similar, we can experiment with providing ``imperfect prefix embeddings'' (smaller than the full prefix) as artefacts to the model to speed up the inference.
We compute the vector similarity between (1) $\text{embd}(w_{:i-1})$ and $\text{embd}(w_{:i})$ and (2) $\text{embd}(w_{:i-1})$ and $\text{embd}(w_{:\text{last}})$ using the the inner product on normalized vectors and show the results in \Cref{fig:embd_sim}.

We limit the context in which we compute the sentence representation in two modes: right crop and left crop, as illustrated in \Cref{tab:sub_feeders}.
Note that the representation is computed strictly from the left context of the token to be predicted.
They correspond to the prefix of the prefix or suffix of the prefix. 

\begin{figure}[htbp]
\centering
\includegraphics[width=0.85\linewidth]{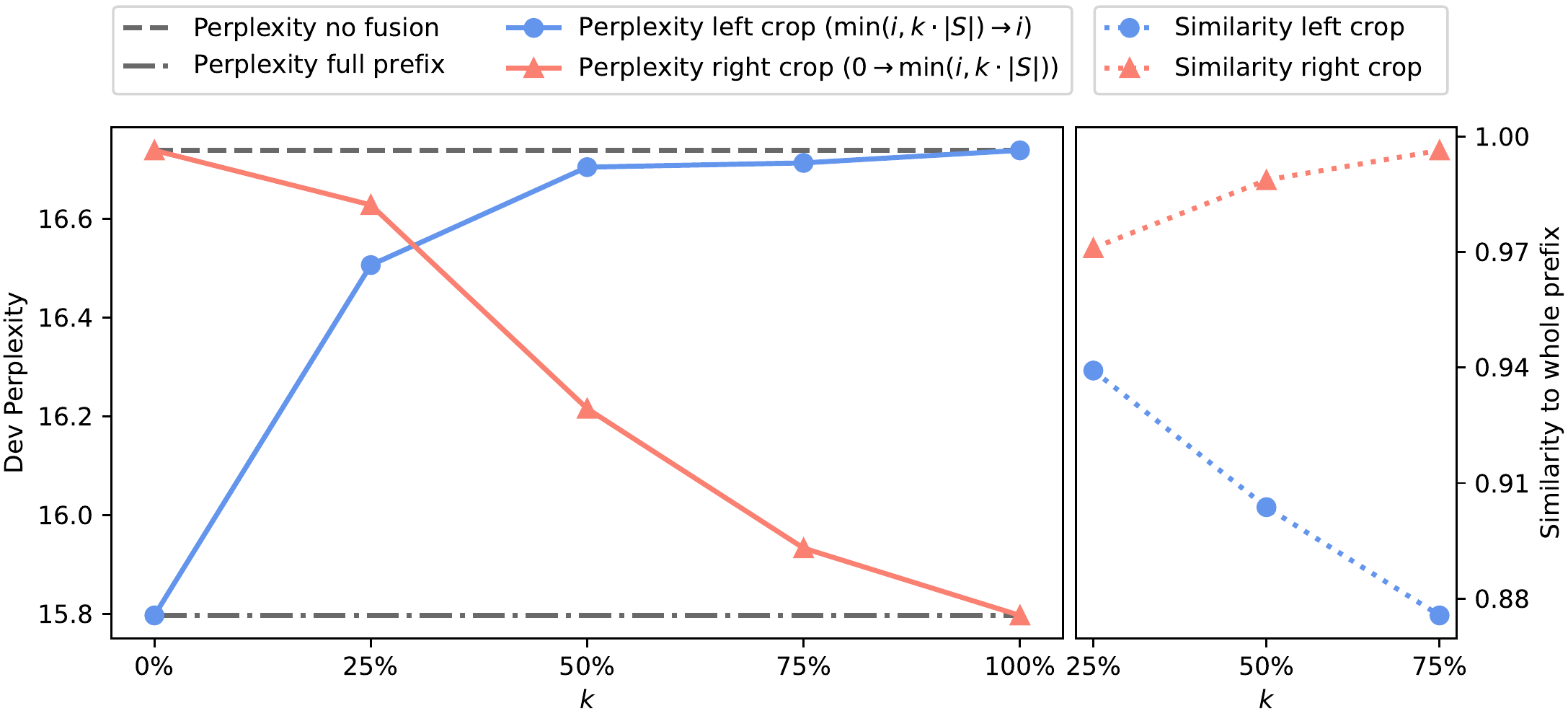}
\caption{Fusion model perplexity with restricted artefact acces (left or right crop) together with average similarity of cropped representations to their full counterparts.
If left \& right crops were equally informative, the graph would be symmetric and the lines would cross in the middle.
Right crop yields lower complexity.
}
\label{fig:sub_feeders}
\end{figure}

Model performance for the various cropped artefact modes is shown in \Cref{fig:sub_feeders}.
Both curves for the left and the right crop (blue and red) are naturally bounded between fusion (bottom) and no fusion (top).
Using the left part of the prefix (right crop) seems to provide more useful information.
This is supported by the vectors being more similar to the full prefix than for the left crop.
It is therefore possible to avoid recomputing the prefix for the last tokens of the sequence at the cost of a slight increase of perplexity.

\citet{khandelwal-etal-2018-sharp} suggest that LSTM LMs model distant past only as a rough semantic field or topic instead of a detailed sequence of tokens.
Having the artefact focus on the distant context (sentence beginning) may therefore provide a more predictive signal in contrast to the immediate context which is still well modelled by LSTM.
It is unclear whether this improvement is caused by the extra semantic information from BERT or by an appropriation of continuos cache or attention architectures.

\begin{table}[htbp]
% hline/vline hacks to add spacing between blocks
% these commands shouldn't be visible beyond this float
\newcommand{\acell}{ \cellcolor{DimGray} }
\newcommand{\bcell}{ \cellcolor{Gainsboro} }
\newcommand{\eol}{ \arrayrulecolor{white} \specialrule{10pt}{0pt}{0pt} }
\newcommand{\eoc}{\color{white}\vline width 1pt}

\resizebox{\linewidth}{!}{%
\renewcommand*{\arraystretch}{0.0}
\begin{tabular}{lc!\eoc c!\eoc c!\eoc c!\eoc c!\eoc c!\eoc c!\eoc c!\eoc c!\eoc c!\eoc c!\eoc c!\eoc}
\toprule 
& & & \footnotesize 25\% & & & \footnotesize 50\% & & & \footnotesize 75\% & $\downarrow$ & & \footnotesize 100\% \\
& Moctezuma & was & the & son & of & the & emperor & Huitzilihuitl & and & \texttt{\_\_\_} & \texttt{\_\_\_} & \texttt{\_\_\_} \\
\cmidrule{2-13}
Right crop 0\% & & & & & & & & & \\ \eol
Right crop 25\% & \acell & \acell & \acell \\ \eol
Right crop 50\% & \acell & \acell & \acell & \acell & \acell & \acell \\ \eol
Right crop 75\% & \acell & \acell & \acell & \acell & \acell & \acell & \acell & \acell & \acell \\ \eol
Right crop 100\% & \acell & \acell & \acell & \acell & \acell & \acell & \acell & \acell & \acell & \bcell & \bcell & \bcell \\ \eol
Left crop 0\% & \acell & \acell & \acell & \acell & \acell & \acell & \acell & \acell & \acell & \bcell & \bcell & \bcell\\ \eol
Left crop 25\% & & & \acell & \acell & \acell & \acell & \acell & \acell & \acell & \bcell & \bcell & \bcell \\ \eol
Left crop 50\% & & & & & & \acell & \acell & \acell & \acell & \bcell & \bcell & \bcell \\ \eol
Left crop 75\% & & & & & & & & & \acell & \bcell & \bcell & \bcell \\ \eol
Left crop 100\% & & & & & & & & & & & & \bcell \\
\arrayrulecolor{black} \bottomrule
\end{tabular}%
}

\caption{Consecutive dark grey rectangles indicate substring of which sentence representation was computed and fused. The representation is not computed beyond the to-be-predicted token ( $\downarrow$ ). Light grey words are cropped because they are not in the left context. Right crop 0\% and left crop 100\% correspond to no fusion while right crop 100\% and left crop 0\% correspond to full fusion without any cropping.}
\label{tab:sub_feeders}
\end{table}

\subsection{Limited Data}
\label{subsec:limited_data}

\begin{figure}[htbp]
\centering
\includegraphics[width=0.8\linewidth]{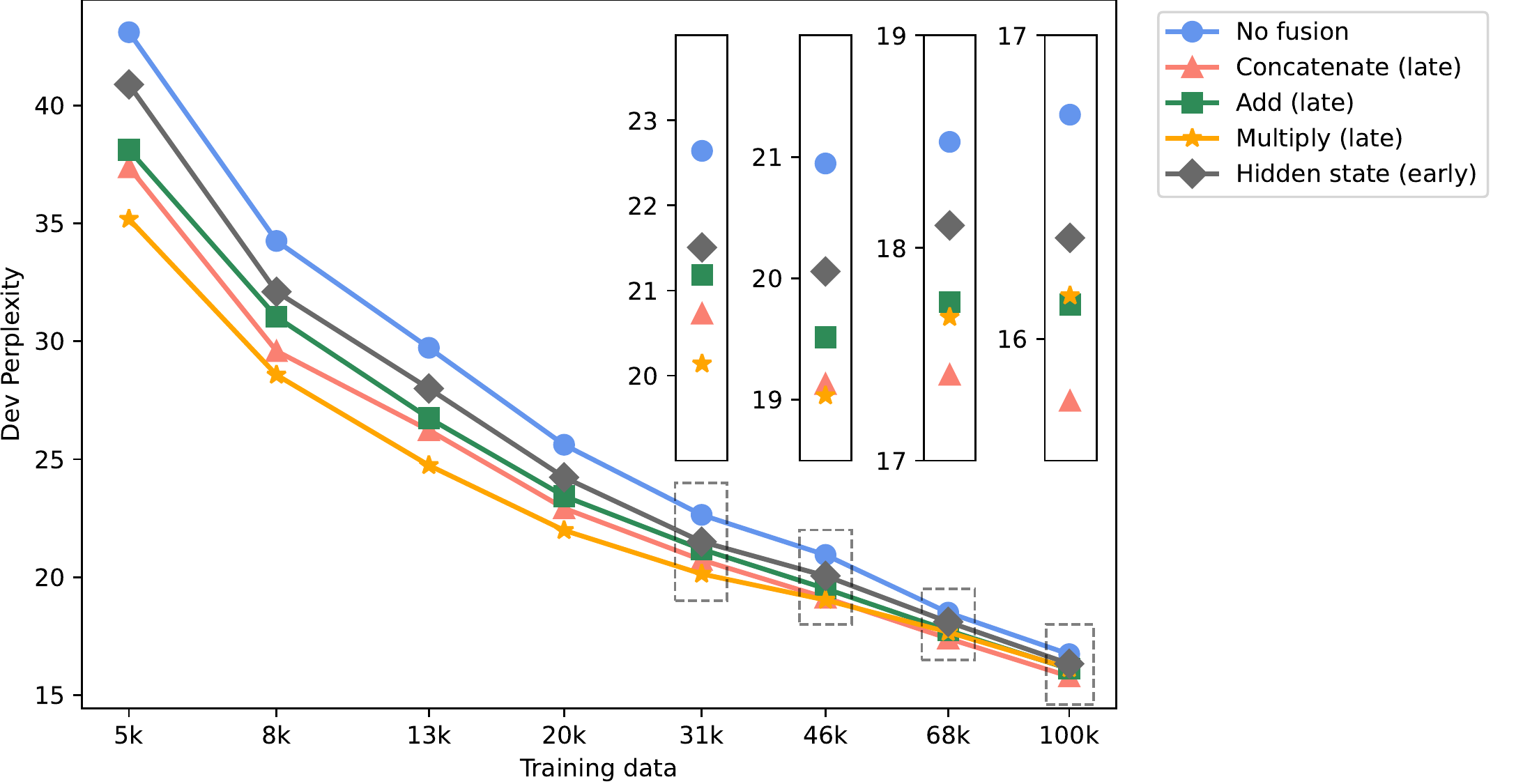}
\caption{Model perplexity evaluated on the same dev dataset with varying training data size (single run). Note the log-scaled x-axis.
Order of fusion performance remains the same for most of the time, with multiplication performance worsening w.r.t. others with more data. 
}
\label{fig:limited_data}
\end{figure}

We further explore whether there is an effect of training data size on the performance of different fusion modes.
While the development set is kept the same, we trim the training data size and report results in \Cref{fig:limited_data}.
While the ordering remains similar (with the exception of multiplication), the differences between the fusion modes become smaller and smaller with increasing amounts of training data.

The benefit of fusion diminishes with more data (in absolute differences).
This finding offers several applications, for example in case of training an autoregressive language model for which there is little training data but there exists a multi-lingual MLM, such as XLM-RoBERTa \citep{conneau2020unsupervised}.

\vspace{-5mm}
\subsection{Wean-off}

\begin{figure}[htbp]
\centering
\includegraphics[width=0.8\linewidth]{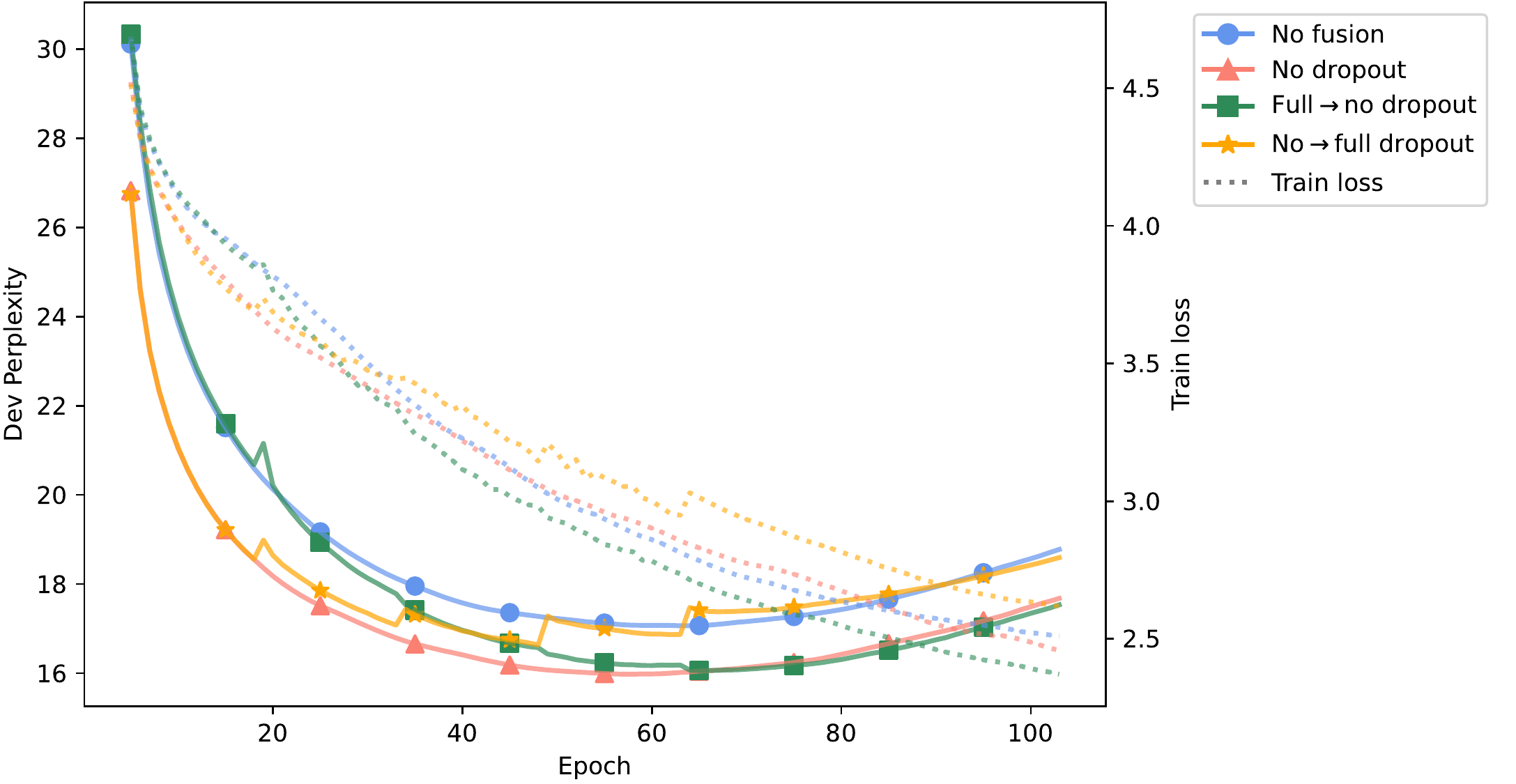}
\caption{Performance during training of ``wean-off'' models with late concatenate fusion. First 5 epochs cropped for increased resolution.}
\label{fig:wean_off}
\end{figure}

Depending on prefix representations computed by a large pre-trained model is not ideal because of the need to run all input through the pre-trained model first. 
%Having a large pre-trained model dependency is not ideal.
We attempted to reduce this dependency by gradually restricting access to the artefacts via dropout.
In \Cref{fig:wean_off} we show (1) a baseline without any fusion, (2) a baseline with fusion and no dropout, (3) a ``wean-off'' model with gradual $0$ to $100\%$ dropout and (4) a reverse model with $100\%$ to $0\%$ artefact dropout.
The dropout for the last two models changes every 15 epochs in the following sequence $(0\%, 25\%, 50\%, 75\%, 100\%)$ or reversed.

The \emph{No $\rightarrow$ full dropout ``wean-off''} model starts the same as the baseline fusion one.
With each change in dropout, however, its performance deteriorates and finally converges to the baseline without any fusion.
The opposite trend can be seen for the  \emph{Full $\rightarrow$ no dropout ``reverse wean-off''} model, which converges close to the no dropout (full fusion) model.
It has a non-smooth transition only with the change of dropout from $0\%$ to $25\%$ because until then, the weights were unactivated and unoptimized.
This shows that the information contained in the artefact is helpful but with the downside that there is no pure optimization benefit to having access to artefacts and it can not be removed.

\subsection{Artefact Source}
\label{subsec:artefact_source}

In this part, we examine the effect of the artefact origin on the autoregressive LM performance.
Specifically, we compare different pretrained LMs that provide the artefact: BERT \citep{devlin2018bert} and SentenceBert \citep{reimers2019sentence}.
The artefact is extracted either as the last layer representation at the \texttt{[CLS]} token or the average of tokens at the last layer.
As shown in the first section of \Cref{tab:artefact_source} the choice of the pre-trained LM and artefact extraction method seems to have a small effect on the actual model performance.
The usefulness of a dense sentence representation is demonstrated by the poor performance of bag of words and TF-IDF baselines.

\paragraph{Full sentence embeddings.}
We additionally compare to fusing the full sentence representations (including the to-be-predicted token) and full sentence representations with the to-be-predicted token masked.
When used as a fusion for the LSTM LM, they both lower the perplexity by a large margin.
However, the full sentence without masking artefact is not applicable during inference because it includes the token that is being predicted.
The masked full-sentence representation can not be used for autoregressive LM because of the dependency on the right context.

% \TODO{It's interesting that bert-cls full sentence makes it better so much (marius surprised). Think about application.}

\begin{table}[htbp]
\center
\begin{tabular}{lcc}
\toprule
& \multicolumn{2}{c}{\hspace{-1cm}\textbf{Model}} \\
\textbf{Artefact source} & \textbf{Fusion} & \textbf{Artefact-only} \\
\midrule
% fustigate, rubinetto
% Shuffled (sentence) & 13.23 & 13.93 \\
% sapid, jersey
% Shuffled (corpus) & 13.89 & 15.31 \\
BERT-CLS & 15.80 & 38.55 \\
BERT-Avg & 15.96 & 48.34 \\
SentenceBert-CLS & 16.08 & 53.01 \\
SentenceBert-Avg & 16.01 & 53.31 \\
\cmidrule{1-1}
% yeowoman
% No fusion & \multicolumn{2}{c}{13.80} \\
Zero artefact (no fusion) & 16.74 & 75.40 \\
Bag of Words & 16.71 & 65.30 \\
TF-IDF & 16.67 & 65.47 \\
\cmidrule{1-1}
BERT-CLS full sentence$^*$ & 12.45 & 55.43  \\
BERT-CLS full sentence, masked$^*$  & 12.52 & 37.87 \\
\bottomrule
\end{tabular}
\caption{Dev perplexity of an autoregressive LM fused with various artefact sources. Artefact sources marked with $*$ can not be used for autoregressive LM because they access right context.}
\label{tab:artefact_source}
\end{table}

\paragraph{Artefact-only model.}
We also attempt to do language modelling based solely on the artefact (and not the currently predicted left context) using a simple non-recurrent feed-forward model described in \Cref{subsec:model}.
This is done in order to measure the amount of information stored in the artefacts alone.
We employ language modelling instead of standard statistical methods (e.g. mutual information) for easy comparison with the full models.
Using the full sentence as an artefact for the artefact-only model performs poorly because the non-recurrent network does not have access to the current position in the sentence.
One input ($x$) to the model is matched with multiple outputs ($y$) based on the sentence length.
It is still better than no fusion because based on the sentence representation, the model can shift the vocabulary distribution to fit the sentence topic.
\Cref{tab:artefact_only_preds} shows the differences in predictions (based only on the artefact) of two sentences.
Similar changes in vocabulary distribution were observed also for the prefix-only model.
Finally, simply masking the currently predicted word position improves the artefact-only model performance substantially.

\begin{table}[htbp]
\center
\begin{tabular}{ll}
\toprule
\textbf{Sentence} & \textbf{Selected top token predictions} \\
\midrule
Moctezuma finally faced the conquistadors. & battle, returned, death \\
The actor finally faced the camera. & story, episode, character \\
\bottomrule
\end{tabular}
\caption{Top non-stopword predictions of the artefact-only model with full-sentence artefacts.}
\label{tab:artefact_only_preds}
\end{table}

The zero-artefact represents no fusion but for fair comparison (same architecture), it is a constant artefact of zeroes.
Training of the artefact-only model with zero artefact means estimating the unigram distribution.
Bag of words and TF-IDF baselines are only slightly better than using no artefact at all.
It is still noteworthy because creating these artefacts is lightweight and does not require any more data than those used for the model training.

\subsection{Cross-Domain}

The sentence representations from BERT are based on a much larger dataset, which could help the LSTM model in cross-domain transfer.
We therefore investigate the effect of fusion on other domains, which were not part of the training set.
The extra datasets, introduced in \Cref{tab:data_size}, are from the book and news domains.
We still train on WikiText but then report results on the two other domains with and without fusion.
\Cref{tab:crossdomain} shows that the gains from fusions still apply even outside of the training domain.

\begin{table}[htbp]
\center
\begin{tabular}{lcccc}
\toprule
& \textbf{WikiText} & \textbf{Books} & \textbf{News} & \textbf{Natural Stories} \\
\midrule
No fusion & 16.74 & 18.10 & 21.52 & 18.10 \\
Fusion (concat.) \hspace{-3mm} & 15.80 & 17.59 & 20.49 & 16.98 \\
\cmidrule{1-1}
\footnotesize Difference/Perc. change & \footnotesize 0.94/5.95\% & \footnotesize 0.51/2.90\% & \footnotesize 1.03/5.02\% & \footnotesize 1.12/6.60\% \\
\bottomrule
\end{tabular}
\caption{Dev perplexity of an autoregressive LM on out-of-domain data with and without fusion (concatenate). Last row shows difference between the two.}
\label{tab:crossdomain}
\end{table}

\section{Surprisal Correlation}

We examine the intuitive hypothesis that better LMs, as measured by perplexity, with richer context representations, will have higher correlations with human surprisal.
We replicate the experiment design by \citet{aurnhammer2019comparing} in measuring the correlation by our LM surprisal with human surprisal proxied via self-paced reading times.
We do not use the same data because of their unavailability.
The model output is turned into a probability-like variable via softmax, however, it is not calibrated.\footnote{The numeric output distribution does not follow the actual distribution of tokens in the text.}
Similarly, the surprisal is measured via the proxy of reading times, which may be a non-linear function that obfuscates the underlying variable, making measuring the actual correlation between surprisals more difficult.
We extract the true class probability \citep{corbiere2019addressing} and, following \citet{wilcox2020predictive}, correlate (Pearson) negative log-likelihoods from the language model with reading times.
We use the mean per-word reading times from Natural Stories \citep{futrell2021natural} with probabilities from a model trained on WikiText (concatenate fusion). 
The probability of a word which is composed of multiple subword units is computed as the product of probabilities (chain rule).

Despite the improvement in perplexity through the fusion, as shown in \Cref{tab:crossdomain}, the correlation with reading times remained largely unaffected: 0.352 (no fusion) $\rightarrow$ 0.349 (concatenate).
Notably, the correlations peaked earlier ($\sim$ epoch 20) than best perplexity ($\sim$ epoch 50), rejecting the original intuitive hypothesis.
This further shows that the tasks of LM and mimicking human surprisal are not identical.

\section{Summary}

In this paper, we presented a small autoregressive LSTM-based LM which makes use of fusion of sentence embeddings provided by a large masked LM.
The exploration of fusion in general can be used for enhancing model performance but also measuring the helpfulness of external information for specific tasks.
We explored multiple fusion modes with late concatenate fusion being better than early or no fusion.

It plays little role which pre-trained LM was used to provide the sentence embeddings.
The prefix vectors were increasingly similar and a tradeoff can be made between speed and lower perplexity by not recomputing the prefix embedding for later tokens in the sentence.
The improved model performance through fusion was preserved even when evaluated on a different domain (books and news instead of Wikipedia text).

This research showed the ease with which extra information, represented as a vector, can be fused into a LM but also that other NLP tasks should be explored from the same perspective as well.

\paragraph{Future work.}

More fusion modes could be explored (e.g. intermediate fusion of token embeddings or embedding of the previous sentence) together with means of speeding up the inference by limiting the number of calls to the masked LM, such as recomputing the artefact only every $k$ steps.
This can be further extended to other architectures (e.g. Transformers) and different artefact sources, such as dense representation of spans retrieved based on the prefix.
% Importantly, this experiment should be replicated in a multi-lingual setting.

An in-depth analysis of the effect of the artefact on the LM inference should be made to determine how exactly it affects the computation.
This can be done e.g. by examining the distances in the intermediate vector spaces between computations with and without artefacts.

Importantly, comparisons with larger baselines should be done, such as GPT-2, to establish the usefulness of this method.
A manual examination of the output could determine whether the improvement in perplexity is due to the disambiguation, as motivated in \Cref{fig:model_moctezuma}.

\paragraph{Limitations.}

A clear limitation of this research is the relatively small model and data size used for experiments, which could be remedied in further studies.
The lack of better correlation of an enhanced LM also documents a limitation to its applications for alignment with some aspects of human reading experience and warrants further investigation.

\section*{Acknowledgements}
This work was funded by the Deutsche Forschungsgemeinschaft (DFG, German Research Foundation) – Project-ID 232722074 – SFB 1102.
We thank the two anonymous reviewers from \href{https://ufal.mff.cuni.cz/pbml}{The Prague Bulletin of Mathematical Linguistics} and its board for their valuable suggestions.

\bibliography{misc/bibliography.bib}

\end{document}